\def\year{2020}\relax
\documentclass[letterpaper]{article} % DO NOT CHANGE THIS
\usepackage{aaai20}  % DO NOT CHANGE THIS
\usepackage{times}  % DO NOT CHANGE THIS
\usepackage{helvet} % DO NOT CHANGE THIS
\usepackage{courier}  % DO NOT CHANGE THIS
\usepackage[hyphens]{url}  % DO NOT CHANGE THIS
\usepackage{graphicx} % DO NOT CHANGE THIS
\usepackage{graphicx}  % DO NOT CHANGE THIS
\usepackage{bm}
\usepackage{amssymb}
\usepackage{booktabs}       % professional-quality tables
\usepackage{pifont}% http://ctan.org/pkg/pifont
\newenvironment{sequation}{\begin{equation}}{\end{equation}}    %  Hi, my dear editor, I am not certain whther this is permitted. 
\usepackage{xcolor}
\title{Simple Pose: Rethinking and Improving  a Bottom-up  Approach \\ for Multi-Person Pose Estimation}
\author{Jia Li\textsuperscript{\rm 1},   Wen Su\textsuperscript{\rm 2}  and   Zengfu Wang\textsuperscript{\rm 3} \\  % \textsuperscript{\rm 1}    All authors must be in the same font size and format. Use \Large and \textbf to achieve this result when breaking a line
\textsuperscript{\rm 1}Department of Automation, University of Science and Technology of China, Hefei, China\\ %If you have multiple authors and multiple affiliations
\textsuperscript{\rm 2}Faculty of Mechanical Engineering and Automation, Zhejiang Sci-Tech University, Hangzhou, China \\
\textsuperscript{\rm 3}Institute of Intelligent Machines, Chinese Academy of Sciences, Hefei, China \\
jialee@mail.ustc.edu.cn, wensu@zstu.edu.cn,  zfwang@ustc.edu.cn \\
}

\begin{document}

\maketitle

\begin{abstract}
 We rethink a well-known   bottom-up approach for multi-person pose estimation  and propose an improved one. The improved approach surpasses the baseline significantly thanks to (1) an intuitional yet more sensible representation, which we refer to as \emph{body parts }to encode the connection information between keypoints, (2) an improved  stacked hourglass network with  attention mechanisms, (3) a novel focal L2 loss which is  dedicated to ``hard’’ keypoint and  keypoint association (body part) mining,  and (4) a robust greedy keypoint assignment algorithm for grouping the detected keypoints into individual poses. Our approach not only works straightforwardly but also outperforms the baseline by  about 15$\%$ in average precision and is  comparable to the state of the art on the MS-COCO test-dev dataset.
 The code and pre-trained models are publicly available online\footnote{\textcolor{magenta}{\url{https://github.com/jialee93/Improved-Body-Parts}}}. 
\end{abstract}

\section{Introduction} \label{introduction}
\label{sec:introduction}

The problem of multi-person pose estimation aims at recognizing and localizing the anatomical keypoints (or body joints) of all persons in a given image. Considerable progress has been made in this field, benefiting from the development of more powerful convolutional neural networks (CNNs), such as ResNet \cite{He2015Deep} and DenseNet \cite{Huang2017Densely},  and more representative benchmarks, such as MS-COCO \cite{Lin2014Microsoft}. 

Existing approaches tackling this problem can be divided into two categories: top-down and bottom-up. The top-down approaches, e.g.   \cite{Chen2017Cascaded,Papandreou2017Towards,sun2019deep},  usually employ a state-of-the-art (SOTA) detector, such as  SSD \cite{Liu2015SSD},   to capture all the persons from the image first.  Then, the cropped persons are resized and fed into the SOTA pose estimator designed for  a single person, e.g.  \cite{Wei2016Convolutional,Newell2016Stacked,Chen2017Cascaded}.
In contrast, the bottom-up approaches, e.g. \cite{Papandreou2018PersonLab,kreiss2019pifpaf}, directly infer  the keypoints  and the connection information between keypoints of all persons in the image without a human detector. Afterwards, the keypoints are grouped to form multiple human poses based on the inferred connection information.

Top-down approaches, e.g. \cite{Papandreou2017Towards,sun2019deep,li2019rethinking}, usually have complicated structures and low performance-cost ratios. Compared with them,  the bottom-up approaches, e.g. \cite{Newell2017Associative,kreiss2019pifpaf},  can be more efficient in inference and independent of human detectors. However, they have to group  the keypoints correctly. And the keypoint grouping (paring or association in other words) can be a big challenge, resulting in another bottleneck for real-time usage  \cite{pishchulin2016deepcut,Iqbal2016Multi}. 
CMU-Pose \cite{Cao2017Realtime}, PersonLab  \cite{Papandreou2018PersonLab}  and PifPaf  \cite{kreiss2019pifpaf} use the greedy parsing algorithm to group detected keypoints into individual poses and break through the bottleneck  to some extent.

It is worth mentioning that the approach proposed by \cite{Cao2017Realtime} (referred to as CMU-Pose here for convenience) is the first  bottom-up approach to perform the task of multi-person pose estimation in the wild with high accuracy,  and almost in real time.  Our approach is mainly inspired by this work but is more  intuitive yet more powerful. Hence, CMU-Pose  is selected to be the baseline approach in this work.

The main contributions of this paper are summarized as follows: (1) we rethink the encoding of joint association, which is named as  \emph{Part Affinity Fields}  (PAFs) \cite{Cao2017Realtime}, and propose a simplified yet more reasonable one, which we call \emph{body parts}, (2) we present an improved stacked hourglass network  with attention mechanisms to generate high-res and  high-quality heatmaps,  (3) we design a novel loss   to help  the network learn ``hard’’ samples, and (4) we develop the greedy keypoint assignment algorithm.

\begin{figure*}[!htp]
	\centerline{\includegraphics[width=1.98\columnwidth]{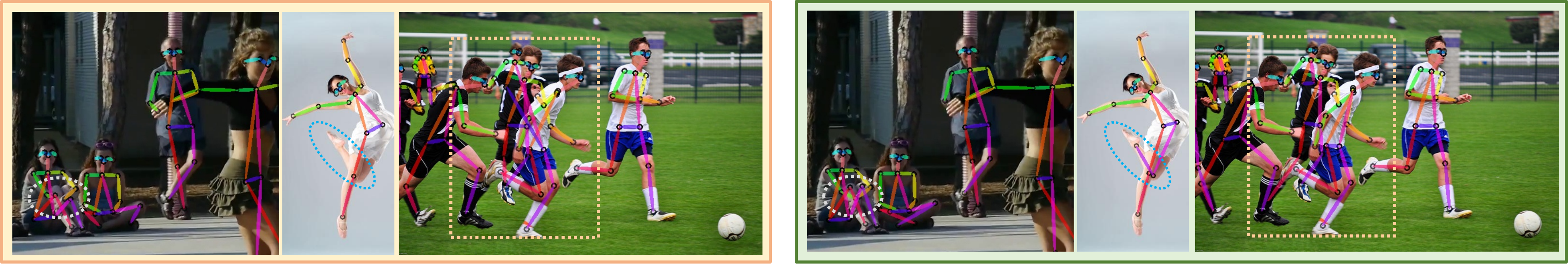}}
	\caption{ Qualitative comparison between our approach and CMU-Pose  \cite{Cao2017Realtime}.
		Left: results produced by CMU-Pose. Right: our results. CMU-Pose  works even when many people appear in the scene, but it suffers precision loss in keypoint localization and it can not detect or group the ``hard'' keypoints (such as the occluded  keypoints) well.   By comparison, our approach is more accurate in keypoint localization and more robust to complex poses and moderate overlaps.}
	\label{fig1}
	%\vspace{-1cm}   %  , thanks to the propose novel loss and improved keypoint assignment algorithm
\end{figure*}

\section{Related Work and Rethinking}

\subsection{Single Person Pose Estimation}  \label{introduction-single-pose}
Classical approaches tackling the problem of person pose estimation  are mainly based on the  pictorial structures  \cite{Fischler1973The,Andriluka2009Pictorial} or the graphical models  \cite{Chen2014Articulated}. They usually formulate this problem as a tree-structured or graphical model problem and detect the keypoints based on hand-crafted features \cite{Chen2017Cascaded}. Recently, the SOTA approaches leverage advanced CNNs and more abundant datasets, making enormous progress in pose estimation. Here, we mainly discuss the CNN based approaches.

DeepPose \cite{Toshev2013DeepPose}  employs  CNNs to solve this problem for the first time, by regressing the Cartesian coordinates of the joints (or keypoints) directly. By contrast,  the work \cite{Tompson2014Joint} presents  CNNs to firstly predict the Gaussian response heatmaps of keypoints, and subsequently it obtains the keypoint positions via finding the local maximums in the heatmaps. Some up to date work, e.g. \cite{Papandreou2018PersonLab,kreiss2019pifpaf},  decomposes the problem of keypoint localization into two subproblems at each pixel location:  (1) binary classification (0 or 1),   and (2) regression of offset vector  to the nearest keypoint.
%Wei et al.  \cite{Wei2016Convolutional} and Newell et al. \cite{Newell2016Stacked} propose multi-stage CNNs with the intermediate supervision training strategy to predict the heatmaps of keypoints from coarse to fine. On the other hand, some SOTA approaches 
%\cite{xiao2018simple, li2019rethinking, sun2019deep}
%based on the single-stage architecture employ very powerful backbone networks such as ResNet-152 \cite{He2015Deep} to infer keypoint heatmaps.

\subsection{Multi-Person Pose Estimation}  \label{introduction-multi-pose}
\subsubsection{Top-down approaches.}

Most of the SOTA results have been achieved by the top-down approaches, such as CPN \cite{Chen2017Cascaded} and HRNet \cite{sun2019deep}. 
Benefiting from the existing well-trained person detectors, the SOTA top-down approaches bypass the difficult subproblem of human body detection and turn the  detection challenges into their advantages.
However, they depend on the human detector heavily and they perform the task in two separate steps. 
The inference time will significantly increase if many people appear together.  

\subsubsection{Bottom-up approaches.}
\label{bottom-up-approaches} The bottom-up approaches, e.g. \cite{Cao2017Realtime,Newell2017Associative,kreiss2019pifpaf}, are more efficient in keypoint inference and do not rely on the human detector. However, they tend to be less accurate. One main reason is that too large and too small persons in the image are difficult to detect at the same time (pose variation and feature map down-sampling  make things even worse). Another main reason lies in the fact that the offset of only a few pixels away from the annotated keypoint location can lead to a big drop in  the evaluation metrics  \cite{wang2018magnify} on the  MS-COCO benchmark \cite{Lin2014Microsoft}. On the contrary,  the top-down approaches are immune to these challenges.

\subsection{Some Rethinking} 
The topic of scale invariance in pose estimation is of great importance. Image pyramid (or multi-scale search in other words)  technique  is usually employed  during testing to cover  the human poses of different scales as much as possible \cite{Cao2017Realtime,Newell2017Associative,Papandreou2018PersonLab}, while  the network is supervised at relatively a smaller scale range during training. Besides, some related work, e.g. \cite{Newell2016Stacked,Chu2017Multi,ke2018multi}, has designed special model structures to enhance the model invariance across scales.

High-res and  high-quality  feature maps (include the output heatmaps) are critical for accurate keypoint localization. Offset regression in PersonLab  \cite{Papandreou2018PersonLab}  and CornerNet \cite{law2018cornernet}, integral pose regression \cite{sun2018integral},   and retaining \cite{sun2019deep} or even magnifying  \cite{wang2018magnify} the resolution of feature maps through the network  are all good tries to relieve  the precision loss (can not be avoided) caused by   image or feature map resizing and small input or feature map size.  

The encoding of connection (or association) information between keypoints is  paid a lot of attention in some prior work, e.g. \cite{Cao2017Realtime,Newell2017Associative,kreiss2019pifpaf,Papandreou2018PersonLab}. New representations bring about new ways of addressing problems.  In this work, we only review the encoding of joint association named \emph{Part Affinity Fields} \cite{Cao2017Realtime} due to the limited space. 

\label{balance}
Another topic worthy of  mention is the problem of imbalanced data: ``positive’’ samples vs ``negative’’ samples (between classes) and ``easy’’ samples vs ``hard’’ samples (within classes). A (Gaussian response) heatmap has most of its area equal to zero (background) and only a small portion of it corresponds to the Gaussian distribution (foreground). Thus, the spread of the Gaussian peaks should be controlled properly to balance the foreground and background. On the other hand, too many easy samples (such as Gaussian peaks of facial keypoints and easy background pixels) can prevent the network from learning the ``hard'' samples (such as Gaussian peaks corresponding to occluded keypoints or body parts) well. These two types of data imbalance problems are critical and  should be addressed properly.

\begin{figure*}[!htp]
	\centerline{\includegraphics[width=1.94\columnwidth]{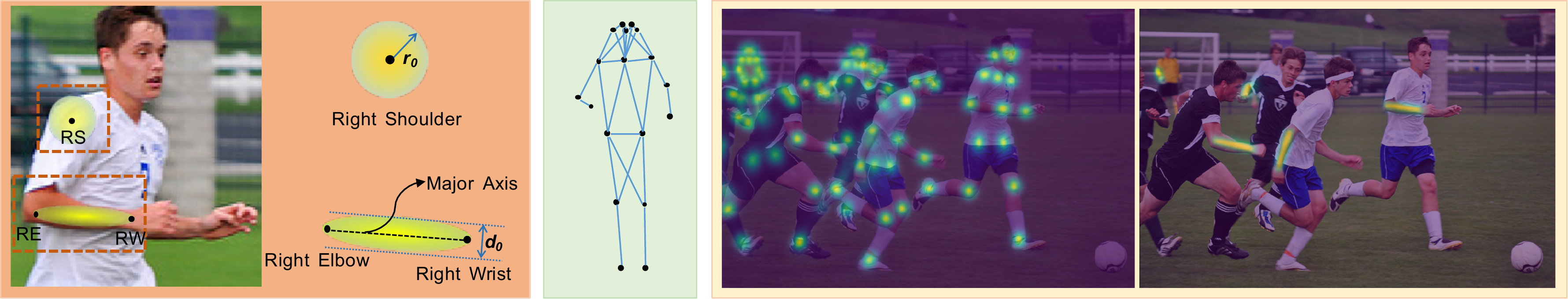}}
	\caption{Definition of heatmaps.  Left:  examples of keypoint Gaussian peak and body part Gaussian peak.
		Middle:  the human skeleton with redundant connections (which we refer to as redundant body parts).  Right: examples of the inferred heatmaps by our network in practice.}  %  The black dots in the  image represent the 2D locations of the annotated keypoints (i.e., right shoulder, right elbow and right wrist respectively). 
	\label{fig2}
	%\vspace{-0.4cm}
\end{figure*}

\section{Our Approach}
We perform the task in three steps: (1) we predict the keypoint heatmaps and the body part heatmaps of all persons in a given image, (2) we get the candidate keypoints and body parts by performing   \emph{Non-Maximum Suppression}  (NMS) on  the inferred  heatmaps, and (3) we perform the keypoint assignment algorithm and collect all individual poses. 

\subsection{Definition of Heatmaps}
Considering the vague concept of keypoints and body parts, and human annotation variances (jitters), our network is supervised to regress the Gaussian responses (0$\sim$1 values) around the keypoint or body part area before obtaining the final localization, introducing the smooth mapping regularization and forcing the network to learn more features nearby.

The heatmaps here include keypoint heatmaps and body part heatmaps.  %The keypoint heatmap encodes the  information of keypoint location.
Each pixel value in the  keypoint heatmaps encodes the confidence that a nearest keypoint of a particular type occurs.  We generate the ground truth keypoint heatmaps by putting  unnormalized  Gaussian distributions with a standard deviation ${{\sigma }_{k}}$ at all annotated keypoint positions.   For example, the generated Gaussian peak of  the right shoulder (RS) keypoint  depicted in the Left of Figure \ref{fig2}. 

% In other words, the ground truth Gaussian distribution at the pixel location ${\bm{p}(x, y)}$ on the ${j}$-th ground truth  keypoint heatmap  is determined by the nearest annotated keypoint of type ${j}$.

The \emph{body part} here refers to the body area which lies between the two adjacent keypoints  (for instance, the forearm area in the Left of Figure \ref{fig2}).  A set of body parts are used to encode the connection information between keypoints  and  extract the visual patterns of human skeleton  (see the Middle of Figure \ref{fig2}). 
Since body part segmentations are not available and  person  masks only cover visible human body areas, we use an elliptical area  to approximately represent the body part.  We generate the ground truth body part heatmaps by putting unnormalized elliptical Gaussian distributions  with a standard deviation ${{\sigma }_{p}}$ in all body part areas.  
% An elliptical Gaussian distribution value at a pixel location ${\bm{p}(x, y)}$  is determined by the shortest vertical distance of this current pixel to all the ellipse major axes. 

By the way, we  map the pixel ${\bm{p}}$ at the location  $(x, y)$ in the ${j}$-th ground truth heatmap   to its   original  floating point location ${\bm{\tilde {  p} }}(\tilde { x }, \tilde { y})={\bm{\tilde {  p} }}(x\cdot R+R/2-0.5, y\cdot R+R/2-0.5)$  in the input image, in which $R$ is the output stride, before generating the precise ground truth Gaussian peaks.

The \emph{PAFs} proposed by \cite{Cao2017Realtime}  is a 2D vector field for each pixel in the  limb area,  which encodes the location and direction information of a limb.   All the pixels within the approximate limb area (may include outliers of the limb) have the same ground truth value, which brings about vagueness or even conflicts  to the information representation.
% In addition, the number of part affinity regression tasks is twice the number of the defined person limbs. 
The  \emph{body part}  representation is more sensible and  composite.  Pixels near to the major axes of the body parts  have higher confidence  and vice versa. 
And we only need the half dimensions of  PAFs to encode the keypoint connection information,  reducing the demand for model capacity. 
During training, a single loss supervises the network to infer two kinds of Gaussian peaks, which have similar  visual patterns and share the same  formulation.

The standard deviations, ${{\sigma }_{k}}$ and ${{\sigma }_{p}}$, control the spread of the Gaussian peaks and they should be set properly to balance  the foreground pixels  and background pixels.  \label{hyperpara}
The hyper-parameters ${{r}_{0}}$ and ${{d}_{0}}$  (see their meanings in Figure \ref{fig2}) determine the boundaries of the ground truth Gaussian  peaks, truncating the unnormalized  Gaussian distribution at a fixed  value $thre$.   It plays a role in our loss function. \label{thre}

\subsection{Network Structure} 

\begin{figure*}[!t]
	\centerline{\includegraphics[width=1.92\columnwidth]{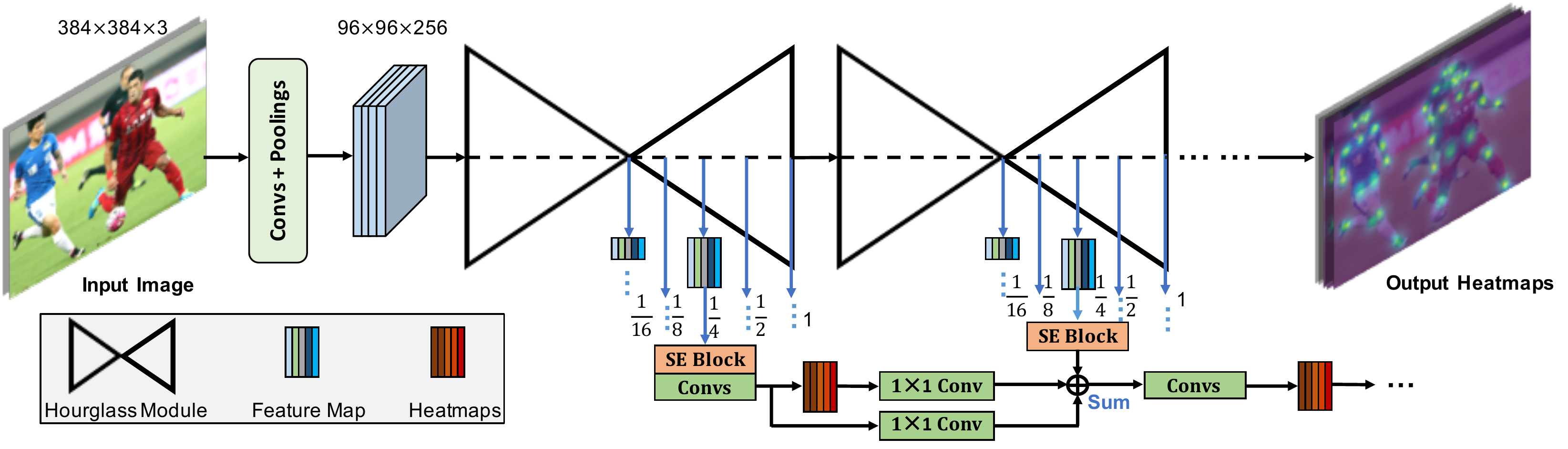}}
	\caption{Identity Mapping Hourglass Network with spatial attention and channel attention mechanisms. 
		% Each convolutional layer in this figure is followed by a batch normalization layer and a LeakyReLu layer, except that the 1$\times$1 convolutional layers are merely followed by batch normalization layers. 
		The feature maps  at 5 different scales (see Figure \ref{fig4}, the feature maps here refer to those surrounded by the blue dashed box in all down-sampling paths) are extracted from each (stage) hourglass module and they are used to produce  heatmaps of different scales. Only the heatmap regression  at the 1/4 scale is illustrated  due to space limitation. The regressed feature maps and heatmaps  from the  previous stage are transformed and reused in the next stage  by element-wise addition (i.e., identity mappings).
	}
	\label{fig3}
	%\vspace{-0.4cm}
\end{figure*}

% Max-pooling operations (non-parametric) and dilated convolutions (parametric) are frequently used in a deep CNN to enlarge the receptive fields.
Large “receptive fields” in CNN  are critical for learning long range spatial relationships and can bring about accuracy improvement \cite{Wei2016Convolutional}.  On the other hand, the detailed information (in smaller receptive fields) is needed for fine-grained localization. 
To consolidate the global and local features, hourglass networks \cite{Newell2016Stacked,Newell2017Associative} have been designed to capture the different spatial extent information of each keypoint and association between keypoints by repeated bottom-up and top-down inference.  In this work, we select the hourglass network, designed for multi-person pose estimation in Associative Embedding (AE) \cite{Newell2017Associative},  as the base model and present an improved one. The improved variant, which we call \emph{Identity Mapping Hourglass Network}  (IMHN), significantly outperforms the   hourglass network in AE \cite{Newell2017Associative}. 

The proposed IMHN,  whose structure  is depicted in Figure \ref{fig3}, is fully convolutional. It takes an image of any shape as the input and outputs multi-scale keypoint  and body part heatmaps of all persons (if any) in the scene simultaneously.
Before fed into the stacked hourglass modules, the original input is down-sampled twofold by some convolutional layers and max- pooling layers.

\begin{figure*}[!t]
	\centerline{\includegraphics[width=1.66\columnwidth]{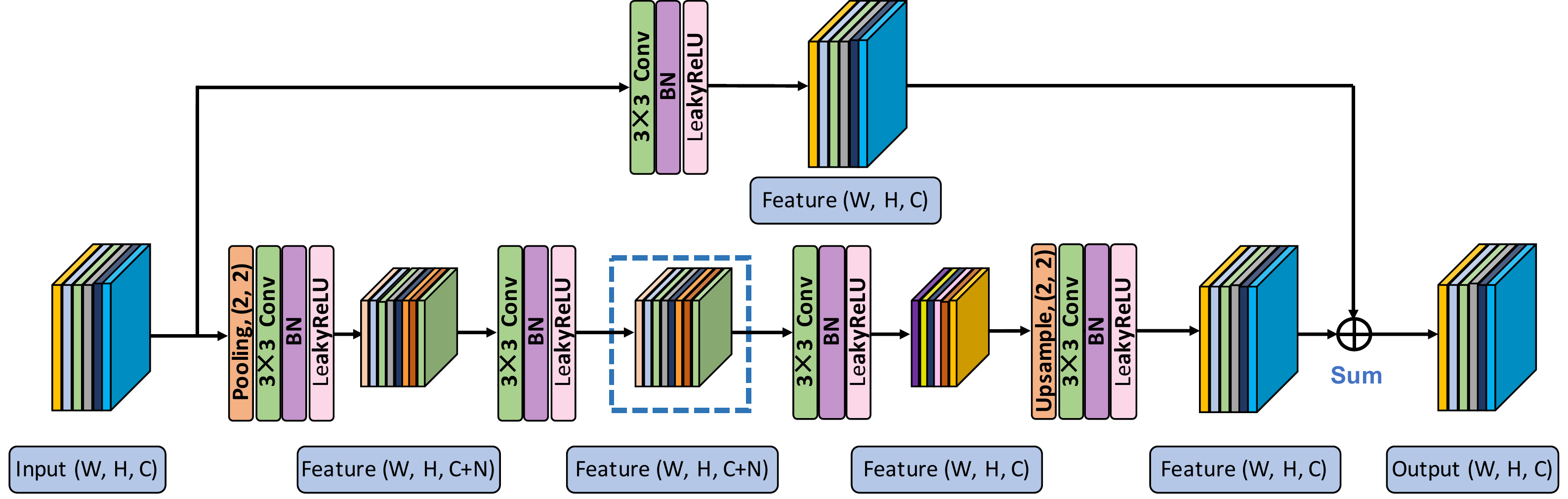}}
	\caption{First order hourglass module. The two branches in this module extract different spatial features and merge them later by element-wise addition.
		%The standard residual connection in ResNet   \cite{He2015Deep}  is used to merge different features from the top and bottom paths and  ease the network training.
	}
	\label{fig4}
	%\vspace{-0.4cm}
\end{figure*}

A first order hourglass module is designed as shown in Figure \ref{fig4}. The down-sampling path reduces the spatial extent of the input feature map by half once and increases the number of  the feature map channels $C$ by $N$ ($C=256$  and $N=128$ in all  experiments unless mentioned otherwise). 
After replacing the dashed box in Figure \ref{fig4} with another first order   module, we  get a second order module.  A  fourth order module can be made by repeating this operation and it is the default hourglass module  to build the IMHN. Therefore, the feature map in the deepest path of our IMHN  has 768 channels.
Please note that we just follow the related work \cite{Newell2016Stacked,Newell2017Associative} to use the fourth order hourglass module and the same $C$ and $N$, ensuring we can make fair comparisons.

During training,  multi-scale supervision \cite{ke2018multi} marked with blue arrows in Figure \ref{fig3} is applied to supervise the fourth order stacked  IMHN  to infer heatmaps at 5 scales from coarse to fine, which   explicitly introduces  the ``spatial attention’’ mechanism. 
Supervising the network at smaller scales can force the network to capture multi-scale structure information of each keypoint and body part.
The low-res heatmaps can provide the guidance of location refinement in the subsequent high-res layers, contributing to the generation of high-quality and high-res heatmaps.
Incidentally, the ground truth heatmaps at fractional scales are down-sampled from the full-size ones using adaptive average-pooling.

The feature map at a certain scale, which is used to regress the heatmaps at the  same scale, is highly self-correlated, for it encodes pose structure information. An SE (squeeze and excitation) block \cite{hu2018squeeze} is inserted into each  feature map at each scale to learn the  channel relationships, which automatically introduces the ``channel attention''  mechanism. Here, we just employ existing techniques to quickly validate   our thoughts.

% the intermediate supervision \cite{Wei2016Convolutional} and
%The introduced ``spatial attention'' and ``channel attention'' mechanism is intuitive but effective according the experiments. 

%We simply implement it using existing modules \cite{ke2018multi,hu2018squeeze}, though some significant insights are revealed. But some different choices and integration can make a big difference

%The SE block \cite{hu2018squeeze} here  introduces the  ``channel attention’’ mechanism to learn the channel relationship within the feature map at each scale. 

Another important innovation in IMHN is that we add identity mappings between the same spatial extent feature maps and heatmaps across different stages (please refer to Figure \ref{fig3}).
%Feature reuse or feature aggregation is frequently  employed in network design, such as in ResNet \cite{He2015Deep} and DenseNet \cite{Huang2017Densely}. 
% We reuse the regressed feature maps and heatmaps  by identity mappings. 
%They transport the predictions produced by the previous stage directly to the final regression layers in the next stage  rather than merely to the initial layer of the next stage. %These shortcuts can enrich the information flow and reuse different scale information across stages without too much cost. 
%According to adequate experiments,  the later stages’ losses in cascaded CNNs with intermediate supervision, e.g. \cite{Wei2016Convolutional,Newell2017Associative}, are obviously  bigger than the first stage’s at the first few training epochs, and they are easy to explode especially in our IMHN. 
They can ease the network training experimentally: stabilizing different stages’ losses and helping the total loss converge faster.

\subsection{Loss Functions}

The L2 loss is frequently used to measure the distance between the predicted heatmaps and the target heatmaps, e.g.  \cite{Wei2016Convolutional,Cao2017Realtime,ke2018multi}. 
To handle the ``hard'' keypoints, the work \cite{Chen2017Cascaded} proposes the L2 loss with online ``hard'' keypoint mining. 
Here, we present a novel loss, which we refer to as \emph{focal L2 loss},	 under the unified definition of keypoint and body part  heatmaps, to deal with the two types of sample imbalance problems as introduced  in Section \textbf{Some Rethinking}.

At each stage of the stacked IMHN, $K$ keypoint heatmaps and $P$ body part heatmaps are inferred at 5 different scales. A pixel value in the inferred heatmaps represents the confidence of being a certain category of keypoint or body part. Assuming the predicted score maps (or heatmaps\footnote{ We use ``heatmap'' and ``score map'' interchangeably throughout our paper for clarity.}) of size $w_{i} \times h_{i}$ at stage $t$ are ${{\bm{S}}^{t}}=\left( \bm{S}_{1}^{t},\bm{S}_{2}^{t},\cdots ,\bm{S}_{K+P}^{t} \right)$, $t\in \left\{ 1,2,\cdots ,T \right\}$, where $T$ is the total number of stacked hourglass modules.
Supposing the ground truth heatmaps of the same size are ${{\bm{S}}^{*}}=\left(\bm{S}_{1}^{*},\bm{S}_{2}^{*},\cdots ,\bm{S}_{K+P}^{*} \right)$ and the Gaussian peak generation function is ${G}$. 
Let $\bm{S}_{j}^{*}(\bm{p})$ denote the ground truth score at the pixel location $\bm{p}(x, y)\in {{\mathbb{R}}^{w_{i}\times h_{i}}}$ in the ${j}$-th  heatmap, we compute $\bm{S}_{j}^{*}(p)$ as: 
\begin{sequation} 
	\bm{S}_{ j }^{ * }\left(\bm{ p} \right) =\left\{ 
	\begin{array}{lr} 
		G({ x,  y  | R,  \sigma  }_{ k },  r_{ 0 }), \quad 1 \le j \le K \\ 
		G({x, y  | R, \sigma  }_{ p }, d_{ 0 }), \quad K < j  \le K+P.
	\end{array} 
	\right.
	\label{eq1}
\end{sequation}
We define $\bm{Sd}_{j}^{t}\left( p \right)$:
\begin{sequation}
	\bm{Sd}_{j}^{t}\left( p \right)=\left\{ 
	\begin{array}{lr}
		\bm{S}_{j}^{t}\left( \bm{p} \right) -\alpha, \:\: \bm{S}_{j}^{*}\left(\bm{ p} \right)>thre \\    %  \hfill right align
		1-\bm{S}_{j}^{t}\left( \bm{p} \right)-\beta , \:\:\: else, 
	\end{array} 
	\right.
	\label{eq2}\end{sequation}
where $thre$  (mentioned in Section \textbf{Definition of Heatmaps}) is the threshold to distinguish between the foreground heatmap pixels and the background heatmap pixels, and $\alpha$, $\beta$ are 
compensation factors to reduce the punishment of easy samples (both easy foreground pixels and easy background pixels) so that  we can make full use of  the training data.  

The focal L2 loss (${\cal FL}$)	 between the predicted heatmaps and target heatmaps of size  $w_{i} \times h_{i}$ at stage $t$  is computed as follow:
\begin{sequation}
	\begin{array}{lr}
		{\cal FL}_{i}^{t} = 
		% \eta \cdot \sum\limits_{j=1}^{K}{\sum\limits_{\bm{p}\in {{\mathbb{R}}^{w_{i}\times h_{i}}}}{\bm{W}\left( \bm{p} \right)\cdot }}\left\| \bm{S}_{j}^{t}\left( \bm{p} \right)-\bm{S}_{j}^{*}\left( \bm{p} \right) \right\|_{2}^{2}\cdot \left\| 1-\bm{Sd}_{j}^{t}\left( \bm{p} \right) \right\|_{2}^{2} \\ 
		\sum\limits_{j=1}^{K+P}{\sum\limits_{\bm{p}\in {{\mathbb{R}}^{w_{i}\times h_{i}}}} \left[ \eta \cdot { \mathbb I } \left(j\le K\right)+1 \right]\cdot  {\bm{W}\left( \bm{p} \right) }} \\
		 \quad \quad \quad \quad  \cdot \left\| \bm{S}_{j}^{t}\left( \bm{p} \right)-\bm{S}_{j}^{*}\left( \bm{p} \right) \right\|_{2}^{2}\cdot \left\| 1-\bm{Sd}_{j}^{t}\left( \bm{p} \right) \right\|_{2}^{2}, \\ 
	\end{array}
	\label{eq3}
\end{sequation}
here, $\bm{W}$  is a binary mask with $\bm{W}\left(\bm{p}\right)=0$ when the annotation is missing at the location $\bm{p}$, $\mathbb I$ is the indicator function, and $\eta$ is the hyper-parameter to balance the keypoint heatmap loss and body part heatmap loss. 
The presented scaling factor term $\left\| 1-\bm{Sd}_{j}^{t}\left(\bm{p} \right) \right\|_{2}^{2}$ implies two prior information: the inferred responses (scores) of easy foregrounds tend to be high (close to 1, e.g.,0.9); the inferred responses of easy backgrounds tend to be low (usually less than 0.01 in practice). Thus, it can automatically down-weight the contribution of easy samples during training, which is inspired by Focal Loss \cite{Lin2017Focal}.

In this work, we set ${{\sigma }_{k}}=9$, ${{\sigma }_{p}}=7$, $thre=0.01$  for the gradient balance between the foreground and background pixels, and we set $\eta=2$ accordingly. In addition, we set $\alpha=0.1$ and $\beta=0.02$ roughly ($\alpha$ and $\beta$ should be set close to $0$ and $\alpha >\beta >0$). For better understandings of the important hyper-parameters, we provide more descriptions.

As for  the standard deviations of keypoint and body part Gaussian peaks, i.e., ${{\sigma }_{k}}$ and ${{\sigma }_{p}}$, if we set them too small, the accurate localization information  is preserved but the inferred responses at these peaks tend to be low, resulting in more false negatives. On the other hand, if we set them too big, the Gaussian peaks spread so flat that the localization information tends to become vague at inference time, harming localization precision (using offset regression may relieve this problem).
As to the hyper-parameter $thre$, it  is set to 0.01  to significantly compress the loss of a mass of easy background pixels. After the network becomes able to distinguish the background well, then, the loss of the foreground starts to play a major role in the network learning.

%The presented scaling factor term $\left\| 1-\bm{Sd}_{j}^{t}\left(\bm{p} \right) \right\|_{2}^{2}$ can automatically down-weight the contribution of easy samples during training, which is inspired  by  Focal Loss \cite{Lin2017Focal}. CornerNet \cite{law2018cornernet} and CenterNet \cite{Zhou:2019ta} also employ focal loss,  but our focal L2 loss is different from theirs both in formulation and function. CornerNet and CenterNet  only regard a single pixel at the  labeled keypoint location as a positive, while we regard  an area around the annotated keypoint as foreground. Our loss works for the regression task of Gaussian distribution   (0$\sim$1), while theirs work for binary classification task (0 or 1). Besides, our loss conveniently does ``hard'' keypoint mining and ``hard'' keypoint association (as the form of body part mining)  meanwhile for the first time.

The total loss of the stacked IMHN across 5 different scales can be written as:
\begin{sequation}
	{ \cal  L }={ \sum _{ t=1 }^{ T }{ \sum _{ i=1 }^{ 5 }{ { { \lambda  }_{ i }^{ t }\cdot  }{ \cal  FL }_{ i }^{ t } }  }  }/{ \sum _{ i=1 }^{ 5 }{ { \lambda  }_{ i }^{ t } }  },
	\label{eq4}\end{sequation}
in which $\lambda_{ 1 }^{ t } =1$, ${{\lambda }_{ 2 }^{ t }}=2$, ${{\lambda }_{ 3 }^{ t }}=4$, ${{\lambda }_{ 4 }^{ t }}=16$ and ${{\lambda }_{ 5 }^{ t }}=64$  are presented  for the balance between  losses at different scales. 
% An improved accuracy  can be expected if we fine-tune the hyper-parameters,  ${{\sigma }_{k}}$,   ${{\sigma }_{p}}$ and so on, according to  keypoint and body part types and person scales \cite{Papandreou2018PersonLab,kreiss2019pifpaf}, but more  hyper-parameters need tuning. 
Now,  ``hard'' keypoints and ``hard'' keypoint association (as the form of  \emph{body parts })    can be  learned  better with the help of the proposed loss under our heatmap definition.

\subsection{Keypoint Assignment Algorithm}   
The candidate keypoints are assigned provided that the candidate body parts are assembled into  corresponding human skeletons. 
We  perform NMS (3 $\times$ 3 window) on the predicted heatmaps to find the candidate keypoints. Then, we obtain the candidate body parts that lie between the candidate adjacent  keypoints,  and calculate their scores that represent the confidence of being body parts,  by sampling a set of Gaussian responses within the body part areas.  
After that, the candidate body parts of the  same type are sorted in descending order according to the weighted scores of body parts and  connected keypoints. 
Consequently, $K$ sets of keypoints $J=\left\{ J{ { s }_{ 1 } },J{ { s }_{ 2 } },\cdots ,J{ { s }_{ K } } \right\} $ and  $P$ sets of body parts $L=\left\{ L{{s}_{1}},L{{s}_{2}},\cdots ,L{{s}_{P}} \right\}$ are obtained. 

Each element ${{l}_{i,j}}\in L{{s}_{i}}$ is a body part instance with type ID ${i}$, connected keypoints and weighted score ${ S }_{l_{ i, j }}$. 
Here, one candidate keypoint can not be shared by two or more body parts of the same type, i.e.,  $\forall$ $j$, $k$ and $m$, $ \left\{ l_{ i,j }, l_{i,k} \right\}  \subseteq  Ls_{i}$ and  $j\ne k$, we ensure that  ${ { l }_{ i,j } }\cap { { l }_{ i,k } }\cap Js_{ m }=\phi$, where $Js_{ m }\in J$. This rule works in analogy to the NMS for candidate bounding boxes in object detection task.
% The reused keypoints and  connected body parts are deleted according to the sorted scores. 

Supposing the  set of assembled human poses is  $H=\left\{ {{h}_{1}},{{h}_{2}},{{h}_{3}},\cdots  \right\}$,  in which ${{h}_{n}}\in H$ represents a single person pose. Then, ${{h}_{n}}$ has $1\sim P$ assigned body parts, corresponding type IDs and total score ${ S}_{{ h}_{n}}$. Our goal is to select the proper candidate body parts and find the best grouping strategy between  the body parts in $ L{{s}_{1}},L{{s}_{2}},\cdots ,L{{s}_{P}} $, such that 
$\sum _{ { h }_{ n }\epsilon H }{ { S}_{{ h}_{n}}} $ reaches to its global maximum.  Instead of solving the problem of global graph matching globally,
 CMU-Pose \cite{Cao2017Realtime} proposes a greedy algorithm upon a minimum spanning tree  (MST) of human skeleton to match the adjacent tree nodes independently   at only a fraction of the  original computational cost.

We follow the greedy strategy  in CMU-Pose  and assemble the human skeletons (see the  Middle of Figure \ref{fig2})  by matching adjacent body parts   independently. 
% (2) the scores and lengths of  body parts, assembled already into human skeletons, are recorded and used as the ``prior knowledge" of pose structure  in the subsequent body part assignments, 
Our keypoint assignment algorithm is based on several simple connection rules. As redundant body parts are introduced in our human skeleton,  the assembled  body parts having lower scores are removed by the body parts having higher scores, which share the same  connected keypoint(s).

\begin{table*}[!ht] 
	\caption{Results on the MS-COCO 2017 validation set.}\smallskip
	\label{table1}
	\centering
	\resizebox{1.96\columnwidth}{!}{
	\smallskip\begin{tabular}{c|c|c|c|c|c|c|c|c|c|c|c}   % 这里面的竖线是小些字母L   {c|p{3.7cm}|c|c|c|c|c|p{3.2cm}|c|c|c|c} 
		\toprule   % %  还可以用\toprule[2pt] 手动改变表格边界线条的粗细
		ID  & \textbf{Method}      	 &Input           &Stride     &${\cal FL}$                   & \textbf{AP}         &ID      &\textbf{ Method}      	 &Input           &Stride      &${\cal FL}$                       & \textbf{AP}   \\   % line 0
		
		\midrule    
		1 &CMU-Pose (6-stage CMU-Net)   	 &368           &8        &N          &56.0       &12 & 3-stage IMHN,  w/ MST           	                       &384          &4       &Y              &65.1      \\   % line 1
		2&AE (4-stage Hourglass, + val data)         	 &512           &4           &N          & 59.7              &13& 4-stage IMHN        	 &384           &4        &N              & 64.5    \\   %  line 2
		3&Top-down$^{*}$  (8-stage Hourglass)    &256      &4      &N             & 66.9              &14&  4-stage IMHN      	 &384           &4         &Y                  & 67.3   \\   %  line 3
		4 &Ours (3-stage CMU-Net)                                  	 &368           &8          &N      & 56.5         &15&  4-stage IMHN, + val data   	 &384          &4        &Y                & 72.3   \\   %  line 5 
		5 &Ours (3-stage CMU-Net)                                           	 &368           &8             &Y      & 60.7         &16& 4-stage IMHN  plus    	 &512         &4       &Y                  & 69.1   \\   %  line 5 
		6 &Ours (4-stage Hourglass)                 &512         &4          &N           &60.0      &17 &  4-stage IMHN  plus, + val data    	 &512           &4        &Y                & 74.1  \\   %  line 4
		7 &3-stage IMHN                                	 &384          &4          &N               &61.5        &18& 4-stage IMHN, one scale       	 &768           &4        &Y                & 63.4   \\   %  line 6
		8 &3-stage IMHN                              	                   &384          &4        &Y             &65.8        &19&  4-stage IMHN plus,  one scale     	 &768          &4      &Y               &65.9   \\   %  line 7
		9 &3-stage IMHN,   w/o  spatial attention                &384          &4        &Y               &64.6        &20 & PifPaf (ResNet-101),   one scale  	 &801         &8      &--                  & 65.7      \\   %  line 7
		10 &3-stage IMHN,   w/o channel attention                     	                   &384          &4        &Y             &65.4      &21&  PersonLab$^{*}$,  one scale &801          &8        &--           &61.2  \\ %  
		11 &3-stage IMHN, ${\cal FL}$  only for keypoint  	 &384         &4      &Y                  &64.2     &22 &  PersonLab$^{*}$,  one scale &1401         &8       &--         &66.5    \\  %8  
		
		\bottomrule  % 可以在未尾行(\bottomrule[2pt])加粗

	\end{tabular}}
\end{table*}

\begin{table*}[!ht] 
	\caption{Results on the MS-COCO 2017 test-dev set. }\smallskip  
	\label{table2}
	\centering
		\resizebox{1.98\columnwidth}{!}{
		\smallskip\begin{tabular}{c|c|c|c|c|c|c|c|c|c|c}   % 这里面的竖线是小些字母L     {p{4.1cm}|c|c|c|c|c|c|c|c|c|c} 
		\toprule   % %  还可以用\toprule[2pt] 手动改变表格边界线条的粗细
		\textbf{ Method}                                                    &Backbone      &Pretrain      &Train Input    &Test Input	  &Refine	& \textbf{AP} 		  &AP$^{M}$	   &AP$^{L}$  	&AR	    &AR$^{50}$	  \\
		\midrule 
		\multicolumn{11}{c}{\textbf{Bottom-up:} multi-person keypoint detection and grouping} \\
		\cmidrule(r){1-11}                   
		CMU-Pose (\textbf{baseline})    \cite{Cao2017Realtime}	                  &6-stage CMU-Net                &N            &368$\times$368        &$\sim$  368$^{2}$            &N  		        &52.9		&50.9	&57.2	&57.0	&79.2 \\ 
		CMU-Pose$^{*}$ \cite{Cao2017Realtime}	                          &6-stage CMU-Net                &N            &368$\times$368        &$\sim$  368$^{2}$            &Y  		        &61.8		&57.1	&68.2	&66.5	&87.2 \\ 
		AE$^{*}$            \cite{Newell2017Associative}                    &   4-stage Hourglass         &N             &512$\times$512       &$\sim$  512$^{2}$           &Y     &65.5	 	&60.6	&72.6	&70.2	&89.5	\\  
		PersonLab$^{*}$ \cite{Papandreou2018PersonLab}     & ResNet-101                        &Y           &801$\times$801        &$\sim$ 1401$^{2}$               &N            &67.8     &63.0        &74.8    &74.5  &92.2		 \\
		PifPaf \cite{kreiss2019pifpaf}                                &ResNet-101                   &Y             &401$\times$401         &$\sim$ 641$^{2}$               &N       &64.9   & 60.6   &71.2    &70.3      &90.2  \\
		PifPaf$^{*}$  \cite{kreiss2019pifpaf}                                &ResNet-152                   &Y             &401$\times$401         &$\sim$ 641$^{2}$               &N       &66.7   & 62.4  &72.9   &72.2    &90.9  \\
		\cmidrule(r){1-11}  
		Ours-1,   w/ ${\cal FL}$        &3-stage CMU-Net             &N                    &368$\times$368      &$\sim$ 368$^{2}$         &N           &59.3     &56.2    &63.8     &63.5   &84.6       \\
		Ours-2,   w/ ${\cal FL}$      &3-stage IMHN                     &N          &384$\times$384      &$\sim$  384$^{2}$                  &N     &65.2     &63.7    &68.5     &69.8   &87.7     \\
		Ours-3,     w/ ${\cal FL}$      &4-stage IMHN                  &N         &384$\times$384      &$\sim$  384$^{2}$                &N      &66.2     &66.4    &66.6     &71.2    &88.6    \\
		Ours-4       (\textbf{final}),    w/ ${\cal FL}$        &4-stage IMHN    plus            &N         &512$\times$512      &$\sim$  512$^{2}$                &N      &\textbf{68.1}      &66.8    &70.5     &72.1   &88.2    \\
		Ours-5,         w/ ${\cal FL}$        &4-stage IMHN    plus            &N         &512$\times$512      &$\sim$  384$^{2}$                &N      &67.6     &64.5    &72.6     &71.3   &87.6   \\
		\cmidrule(r){1-11}  
		\multicolumn{10}{c}{\textbf{Top-down:} human detection and single-person keypoint detection} \\
		\cmidrule(r){1-11}  
		G-RMI$^{*}$  \cite{Papandreou2017Towards}	      &RseNet-101          &Y         &353$\times$257      &353$\times$257      &--        &64.9      &62.3    &70.0     &69.7    &88.7\\
		% Integral Pose Regression \cite{sun2018integral}      &RseNet-101          &Y         &256$\times$256      &256$\times$256      &--        &67.8      &63.9   &74.0     &--    &--\\
		CPN$^{*}$  \cite{Chen2017Cascaded}                      &ResNet-Inception  &Y        &384$\times$288                &384$\times$288     &--   &72.1         &68.7   &77.2   &78.5   &95.1  \\
		HRNet-W48$^{*}$    \cite{sun2019deep}                                       &HRNet-W48         &Y       &384$\times$288           &384$\times$288        &--     &75.5   &71.9  &81.5  &80.5   &95.7  \\
		\bottomrule  % 可以在未尾行(\bottomrule[2pt])加粗
		
	\end{tabular}}
\end{table*}

\section{Experiments}

\subsection{Implementation Details}

\subsubsection{Dataset and evaluation metrics.}

Our models are trained and evaluated on the MS-COCO  dataset \cite{Lin2014Microsoft},
%\footnote{MS-COCO dataset is  available at \url{http://cocodataset.org}.}  
which consists of  the training set (includes around 60K images), the test-dev set (includes around 20K images) and the validation set (includes 5K images). The MS-COCO evaluation metrics,  OKS-based\footnote{ OKS (Object Keypoint Similarity) defines the similarity between different human poses. Only the top 20 scoring poses are considered during evaluation.} average precision (AP) and average recall (AR), are used to evaluate the results. 

\subsubsection{Training details.}

The  training images with random transformations are cropped and resized to the fixed spatial extent of 384 $\times$ 384. And the generated ground truth heatmaps have the  size of  96 $\times$ 96.
We implement both the 3-stage IMHN and 4-stage IMHN using Pytorch.  To train the networks, we use the SGD optimizer with the learning rate of 1e-4 (multiplied by 0.2 for every 15 epochs), the momentum of 0.9, the batch size of 32 and the weight decay of 5e-4.  We train the  IMHNs  with L2 loss first and then continue to train them with the focal L2 loss  (${\cal FL}$)	 until the performance refuses to improve. 
%A warm-up learning rate schedule is used at the first 3 epochs to stabilize training.  
The last but not least, our networks are in mixed precision
%  \footnote{Automatic Mixed Precision Package is publicly available at  \url{https://github.com/NVIDIA/apex}.} 
 to reduce the memory consumption and   speed up the experiments.  The implemented 4-stage IMHN (see Ours-3 in Table \ref{table2}) can be trained at the speed of 33 FPS with 4 RTX 2080TI GPUs, and we train it for about 3 days.

%$\times$0.5, $\times$1, $\times$1.5, $\times$2, $\times$4
\subsubsection{Testing details.}
We first resize and pad the input image so that it fits the network.  Then, the  input  image is inferred at multiple scales (for example, $\times$0.5, $\times$1, $\times$1.5, $\times$2) with flip augmentation. Next, the inferred heatmaps are averaged across scales (i.e., multi-scale search).  
After collecting the poses from the  heatmaps via the keypoint assignment algorithm, we sort them in descending order according to their scores in heatmaps. To compare equally, we have run the released models of CMU-Pose \cite{Cao2017Realtime},  AE \cite{Newell2017Associative} and PifPaf \cite{kreiss2019pifpaf}, obtaining their results without the refinement for detected person poses.

\subsection{Results and Analyses}
\subsubsection{Results on the MS-COCO dataset.}

% Firstly, we only train the corresponding networks on the training set  to compare  with other approaches in Table \ref{table2}  on the test-dev set equally.    Then, we continue to train them on both the training set and  the validation set (leads to  the overfitting on the training  set and validation set  in Table \ref{table1})  to explore the lifting space of our approach. 
We have trained all our networks from  scratch only with the MS-COCO data. The hyper-parameters  of our approach are tuned according to the performance on the validation set.  We compare our approach with the state of the art in Tables \ref{table1} and \ref{table2}.  The backbone of 4-stage IMHN has nearly the same number of convolutional layers as ResNet-101 \cite{He2015Deep}. Thus, we specially compare our networks with those  based on ResNet-101 backbone in  PersonLab  \cite{Papandreou2018PersonLab}  and PifPaf \cite{kreiss2019pifpaf} impartially. The entries  marked with ``*'' in Tables \ref{table1} and \ref{table2} are results reported in their original papers.

The notations in our tables are explained as follows:  ``Input'' denotes the long edge of the test image, ``Stride'' is the ratio of the input image size to the output feature map size, 
% ``$S_{thre}$'' is the threshold to filter the background Gaussian responses, 
``MST'' is short for minimum spanning tree of human skeleton without redundant body parts, ``CMU-Net'' represents the cascaded CNN used in  CMU-Pose \cite{Cao2017Realtime}, ``Hourglass'' denotes the hourglass network, ``Refine'' indicates whether or not  the result is  additionally refined by a single person pose estimator. 
 % And ``w/ all'' means all techniques we proposed are used,  unless mentioned otherwise.  
 The entry named ``Top-down'' in Table  \ref{table1}  is a top-down approach cited from CPN \cite{Chen2017Cascaded}. It employs a SOTA human detector and an 8-stage hourglass network for single person pose estimation.  

\subsubsection{Inference speed.}
The speed of our system is tested on the MS-COCO test-dev set  \cite{Lin2014Microsoft} .
\begin{itemize}	
	\item Inference speed of our 4-stage IMHN with 512 $\times$ 512 input on one 2080TI GPU: 38.5 FPS (100$\% $ GPU-Util).
	\item Processing speed of the keypoint assignment algorithm that is implemented in pure Python and a single process on CPU: 5.2 FPS (has not been well accelerated).
\end{itemize}

\subsubsection{Ablation studies.}
Some examples  of the  qualitative comparison between our approach and CMU-Pose is illustrated in Figure \ref{fig1}. 
The detailed ablation experiments of our approach, numbered  for  clarity,  are  shown in Table \ref{table1}.  Experiments 1 and 4  use different encodings of  keypoint  association  (\emph{PAFs} and \emph{body parts} respectively). They reveal that the encoding of  \emph{body parts}   is better.  According to experiments 2 and 6,  our approach equipped with the same network and L2 loss is comparable to AE which utilizes validation data.   The focal L2 loss can bring about a \textbf{3$\sim$4\bm{\%}}  AP improvement over the L2 loss   (see experiment 5 vs 4, experiment 8 vs 7 and experiment 14 vs 13) under the same definition of heatmaps.  Experiments 8 and 11 demonstrate doing ``hard'' keypoint and body part (keypoint association) mining meanwhile is better.  The  3-stage IMHN can lead to around a \textbf{5\bm{\%}}  AP improvement  compared with the 3-stage CMU-Net (see experiment 7 vs 4 and experiment 8 vs 5), and the 4-stage IMHN  outperforms the 4-stage hourglass network (see experiment  13 vs 6) by a big margin,  indicating IMNHs' advantages.

 The introduced ``spatial attention'' and ``channel attention''   mechanisms  contribute 1.2$\%$  and 0.4$\%$ AP increase respectively to the 3-stage IMHN (see experiments 8, 9 and 10).  The  redundant connections in human skeleton bring about a 0.7$\%$ AP improvement according to the comparative experiments of 8 and 12. Bigger input size or feature map size and more stacks can improve the accuracy consistently according to the results in Table \ref{table1}. Thus, we continue to train the model  in experiment 14 with 512 $\times$ 512  input and more data augmentation,  and obtain the model named ``4-stage IMHN plus'' in experiment 16.  Further, the 4-stage IMHN is able to fit the MS-COCO train-val data at 74.1$\%$ AP (see experiment 17), indicating the big promotion space of our system.

\subsubsection{Comparisons with the state of the art.}

All the models compared in the tables are evaluated without model ensemble. %, except that PersonLab    and PifPaf   have  additional  employed the technique of model parameter averaging. 
Some SOTA top-down approaches may outperform all the SOTA bottom-up approaches including ours, but they all depend on advanced human detectors and use very powerful networks for single person pose estimation.
According to the results,  it is safe to conclude that our approach outperforms the baseline by a big margin (experiment 5 vs 1 in Table \ref{table1} and Ours-1 vs the baseline in Table \ref{table2}) and even surpasses the TOP-DOWN approaches with equal level  backbones (please refer to ``Top-down''  in Table  \ref{table1}  and  ``G-RMI'' in Table  \ref{table2}). 

%Our IMHNs are more robust to scale variation   and  can generate high-quality heatmaps, leading to the precision improvement (see 4-HG w/o ${\cal FL}$ vs   3-IMHN     w/o    ${\cal FL}$ in  Table \ref{table1}). 

Our approach has achieved comparable or even better results on both single scales (see the experiments  with IDs 18$\sim$ 22 in Table \ref{table1}) and multiple scales (see  the results in Table \ref{table2}), compared with the latest SOTA bottom-up approaches, PifPaf and PersonLab, under fair conditions. However, PersonLab benefits greatly from the big input size (see experiment 22 vs 21 in Table \ref{table1}).  
It can be seen that PifPaf and Personlab are superior to our approach in keypoint average recall (AR). But our approach is superior to them when it comes to the robustness to person scales. Both PersonLab and PifPaf drop over 10$\%$ from AP$^{L}$ metric to AP$^{M}$ metric (see the entries in Table \ref{table2}), while our approach (Ours-4) performs well  regardless of the person scales, be they middle scales (M) and large scales (L).
What is more, most of other work benefits greatly  from  the  networks pre-trained for the ImageNet classification task \cite{russakovsky2015imagenet}, while we train  networks from scratch (see experiment 17). 

Referring to the results on the MS-COCO test-dev set  in Table  \ref{table2}, the proposed techniques except for the designed IMHN bring about a \textbf{6.4\bm\%}  AP improvement (Ours-1 vs baseline). The  3-stage IMHN can lead to a 5.9$\%$ AP increase compared with the 3-stage CMU-Net (Ours-2 vs Ours-1).  The  4-stage IMHN can further contribute  a 1$\%$ AP improvement  over the 3-stage IMHN. 
And our final model, ``4-stage IMHN plus'' with bigger input (means bigger output feature maps), brings about a \textbf{15.1\bm\%}  AP improvement in total over the baseline. Incidentally, our approach significantly  outperforms CMU-Pose and AE, though they  have additionally  refined the results using a single person pose estimator.

\section{Conclusions}    
In this paper, we rethink and  develop a bottom-up approach for multi-person pose estimation.  We provide some insights into valuable design choices:
 (1) doing hard sample mining of keypoint and keypoint association meanwhile, (2) using a powerful network to generate high-res and high-quality heatmaps, and (3) introducing the scale invariance across person scales, which  are more critical to improve the performance. 
The experimental results have demonstrated the significant improvement achieved by our approach over the  baseline  (+15.1$\%$ AP on the MS-COCO test-dev dataset).
To the best of our knowledge, our approach, which  is straightforward  and easy to follow, is the first bottom-up approach to  provide both the  source code and pre-trained models with  over 67$\%$ AP on the MS-COCO test-dev dataset.

\section{Acknowledgment}

Special thanks to Assoc. Prof. Qiongru Zheng who helped check and correct language mistakes.

{\small
	\bibliographystyle{./aaai}
	\bibliography{2987.mybib}
}

\end{document}